\documentclass[11pt,a4paper]{article}
\usepackage{coling2020}
\usepackage{times}
\usepackage{url}
\usepackage{latexsym}
\usepackage{listings}
\usepackage{booktabs}
\usepackage{url}
\usepackage{graphicx}
\usepackage{appendix}
\usepackage{multicol}
\usepackage{caption}
\usepackage{subcaption}
\usepackage[ruled, linesnumbered]{algorithm2e}
\usepackage{amsmath}
\usepackage{amssymb}
\usepackage{float}
\usepackage{multirow}
\usepackage{multicol}
\usepackage{xcolor}
\usepackage{makecell}

\colingfinalcopy 

\title{Towards Accurate and Consistent Evaluation: A Dataset for Distantly-Supervised Relation Extraction}

\author{
	Tong Zhu$^1$, Haitao Wang$^1$, Junjie Yu$^1$, Xiabing Zhou$^1$, \\ 
	{\bf Wenliang Chen$^1$, Wei Zhang$^2$, Min Zhang$^1$} \\
	$^1$Institute of Artificial Intelligence, School of Computer Science and Technology, \\ Soochow University, China\\
	$^2$Alibaba Group, China \\
	{\tt \{tzhu7, htwang2019, jjyu\}@stu.suda.edu.cn } \\
	{\tt \{zhouxiabing, wlchen, minzhang\}@suda.edu.cn} \\
	{\tt lantu.zw@alibaba-inc.com}
}

\date{}

\begin{document}
\maketitle

\begin{abstract}
In recent years, distantly-supervised relation extraction has achieved a certain success by using deep neural networks. Distant Supervision (DS) can automatically generate large-scale annotated data by aligning entity pairs from Knowledge Bases (KB) to sentences. However, these DS-generated datasets inevitably have wrong labels that result in incorrect evaluation scores during testing, which may mislead the researchers.
To solve this problem, we build a new dataset NYT-H, where we use the DS-generated data as training data and hire annotators to label test data. 
Compared with the previous datasets, NYT-H has a much larger test set and then we can perform more accurate and consistent evaluation.
Finally, we present the experimental results of several widely used systems on NYT-H.
The experimental results show that the ranking lists of the comparison systems on the DS-labelled test data and human-annotated test data are different.
This indicates that our human-annotated data is necessary for evaluation of distantly-supervised relation extraction.
\end{abstract}
\section{Introduction}
\label{section:introduction}

There has been significant progress on Relation Extraction (RE) in recent years using models based on machine learning algorithms \cite{mintz2009distant,hoffmann2011multir,zeng2015pcnn,zhou2016attblstm,ji2017distant,su2018exploring,qin2018rl,zhang2019capsule}. The task of RE is to identify semantic relationships among entities from texts.
Traditional supervised methods require a massive amount of annotated data, which are often labelled by human annotators. However, it is hard to annotate data within strict time limits and hiring the annotators is non-scalable and costly.

\begin{figure}[ht]
    \centering
    \includegraphics[scale=0.4]{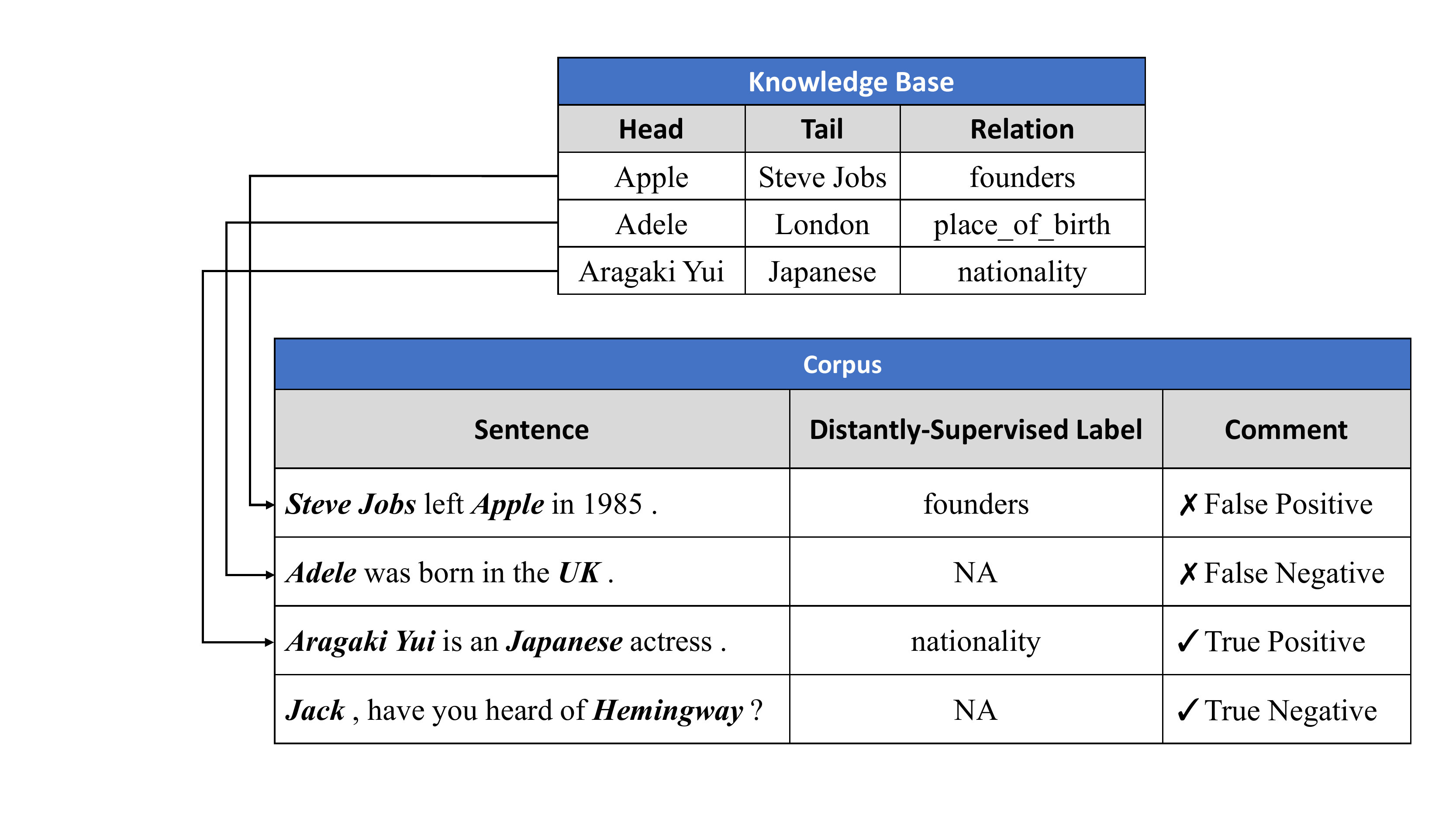}
    \caption{Distant supervision alignment and noises.}
    \label{fig:ds_alignment}
\end{figure}

In order to quickly obtain new annotated data, Mintz et al. \shortcite{mintz2009distant} use Distant Supervision (DS) to automatically generate labelled data. Given an entity pair and its relation from a Knowledge Base (KB) such as Freebase, they simply tag all sentences containing these two entities by this relation. 
Through this way, the framework has achieved great success and brought state-of-the-art performance in RE \cite{qin2018dsgan,liu2018-neural-relation,feng2018reinforcement,chen2019uncover}. 
But as an exchange, the auto-generated data may be of lower quality than those from annotators. 
For example, the sentence ``\emph{Steve Jobs} left \emph{Apple} in 1985 ." is not suitable for relation \emph{founders} labelled by the 
triple $<$\emph{Steve Jobs}, founders, \emph{Apple}$>$ in knowledge base. 

To address the above problem, previous studies regard distantly-supervised relation extraction as a Multi-Instance Learning (MIL) task to reduce the effect of noises while training \cite{hoffmann2011multir,surdeanu2012miml,zeng2015pcnn,lin2016pcnnatt,han2018hnre}.
MIL aggregates sentences into bags, where instances with the same entity pair are grouped into one bag.
However, the wrong labelling problem still remains during system evaluation. There are two types of noises: False Positive (FP) means that a sentence is tagged with a relation when it is not, and False Negative (FN) means that a sentence is tagged with NA relation due to the KB incompleteness, where NA relation means there is not an available relation for the sentence or the relation is not included in the pre-defined relation candidate set.
Figure \ref{fig:ds_alignment} shows the examples of FP and FN. 
Therefore, the performance evaluated on the DS test data might mislead the researchers since the evaluation scores are not exactly correct.

In this paper, we present a new RE dataset NYT-H, which is an enhanced distantly-supervised relation extraction dataset with human annotation based on NYT10 \cite{riedel2010modeling}. 
NYT10 is a widely used benchmark data on the task of distantly-supervised relation extraction, which has been used in many recent studies \cite{lin2016pcnnatt,liu-etal-2017-soft,qin2018rl,qin2018dsgan,chen2019uncover}.
However, the NYT10-test set is also distantly-supervised, which contains many FP and FN noises that may hinder the evaluation.
Besides, NYT10 is not consistent for building classifiers, where the relations in NYT10-train set and test set are not totally overlapped.
There are 56 relations existed in the train set while only 32 relations appear in the test set.
Both the inaccuracy and the inconsistency make the evaluation results less convincing and less objective, which limit the development of distantly-supervised RE.
In NYT-H, we hire several annotators to label test data. With this new data, we are able to perform accurate and consistent evaluation that may boost the research on distantly-supervised RE. Our contributions can be summarised as below:

\begin{itemize}
    \item We present a new dataset NYT-H for distantly-supervised relation extraction, in which we use DS-labelled training data and hire several annotators to label test data. With this newly built data, we are able to perform accurate and consistent evaluation for the task of distantly-supervised relation extraction.
    \item NYT-H can serve as a benchmark of distantly-supervised relation extraction. We design three different evaluation tracks: Bag2Bag, Bag2Sent, and Sent2Sent, and present the evaluation results of several widely used systems for comparison.
    \item We analyse and discuss the results on different evaluation metrics. We find the ranking lists of the comparison systems are different on the DS-labelled and manually-annotated test sets. This indicates that the manually-annotated test data is necessary for evaluation.
\end{itemize}

\section{Related Work}\label{related_work}
\textbf{Distantly-Supervised Data Construction: }Considering the cost and limitations in fully annotated datasets, Mintz et al. \shortcite{mintz2009distant} think all sentences with the same entity pair express the same relation. Based on this assumption, they construct a distantly-supervised RE dataset by aligning Wikipedia articles with Freebase.
Restricted by the Wikipedia corpus, expressions of sentences are very limited in each relation. 
Riedel et al. \shortcite{riedel2010modeling} publish the NYT10, a distantly-supervised dataset which takes the New York Times news as the source texts. 
It is smaller in scale, but contains a variety of expressions. 
Besides, they make a relaxation on Mintz's assumption, they think that at least one instance, which has the same entity pair, can express the same relation.
However, NYT10 does not match this assumption totally. 
Without human annotation, they cannot ensure that there is a correct-labelled sentence in each bag.
Jat et al. \shortcite{jat2018improving} develop a new dataset called Google Distant Supervision (GDS) based on Google Relation Extraction corpus\footnote{\url{https://ai.googleblog.com/2013/04/50000-lessons-on-how-to-read-relation.html}}, which follows Riedel's assumption and makes the automatic evaluation more reliable \cite{vashishth2018reside}.
However, noises that come with DS cannot be eliminated in the test set, so the automatic evaluation cannot be exactly accurate. 

\paragraph{Evaluation Strategy for Distantly-Supervised Relation Extraction: }
There are two measures to evaluate models trained on DS-constructed datasets. 
The first one is the Precision-Recall Curve (PRC), and its Area Under Curve (AUC). 
DS-constructed datasets are usually unbalanced, and PRC is a good way to measure the classification performance, especially when the dataset is unbalanced. 
But PRC does not deal with the DS noises. If we do not have the ground truth answers, the evaluation results will be less convincing and objective. 
The second one is Precision@K (P@K). 
Researchers manually annotate systems' top-K output results \cite{riedel2010modeling,zeng2015pcnn,lin2016pcnnatt} to give relatively objective evaluation results with less cost.
However, there exist annotation biases since the criterions may differ from researchers, and researchers have to annotate different data if the outputs are changed.
Besides, the predicted results are often biased due to the dataset unbalanced problem. This means the top-K instances could not cover all the relations, 
unless the `K' value is large enough.

Noises in DS datasets can provide a simulation to real scenarios, but they are harmful to model selection and evaluation if there are no ground thruths in the test set. So we develop NYT-H, a distantly-supervised dataset with manually annotated test set. We believe this would help researchers to obtain more accurate evaluation results easily.
\section{NYT-H Dataset}\label{sec:data}
In this section, we introduce the procedure of constructing the NYT-H dataset and list the statistics information of the data.
We also compare our dataset with the previous datasets in detail.

\subsection{Data Preprocessing}
NYT-H is built on NYT10\footnote{\url{http://iesl.cs.umass.edu/riedel/ecml/}} \cite{riedel2010modeling}.
There are many data files in the original NYT10, organised in protocol buffer format. 
We use the protobuf\footnote{\url{https://developers.google.com/protocol-buffers}} tool to extract relations and entities.
There are many repetitions between train and test sets. 
Thus a deduplication operation is applied on NYT10 to eliminate duplicates. 
In addition, to enlarge the size of test set, we randomly select some non-NA bags from the NYT10-train set\footnote{We use NYT10-train/test to denote the sets of NYT10, otherwise the train/test sets refer to the sets of NYT-H without explicit mention.}.
Different from NYT10, NYT-H has a NA set that contains all the NA instances. Finally, we obtain three sets: train set, test set, and NA set.

\subsection{Human Annotation}
Then, we hire annotators to label sentences in the test set. The sentences are annotated manually in a binary strategy, which is to decide whether the sentences express the relations assigned by DS. 
It is difficult to pick out one from over 50 types of relations, so we decide to use the binary strategy. 
If a sentence actually expresses the DS-annotated relation for the given entity pair, the sentence will be labelled as ``Yes'', otherwise ``No''. Some examples can be found in Table \ref{tab:annotation_example}, where FP noises (like the first example) will be recognised and annotated as ``No''.

\begin{table}[h]
    \centering
    \scalebox{0.7}{
        \begin{tabular}{c|c|l}
            \toprule
            Annotation&DS Relation&Sentence\\
            \midrule
            No&founders&\textbf{\textit{Steve Jobs}} left \textbf{\textit{Apple}} in 1985 .\\
            Yes&nationality&\textbf{\textit{Aragaki Yui}} is an \textbf{\textit{Japanese}} actress .\\
            \bottomrule
        \end{tabular}
    }
    \caption{Annotation examples. Entities are in bold italic fonts.}
    \label{tab:annotation_example}
\end{table}

There are three annotators working on this project. 
Each sentence is assigned to two annotators. 
If their annotations are different, the third annotator will make the final decision. 
The Kappa coefficient of annotation results is 0.753, which shows a substantial agreement among the annotators. 
Finally, we obtain 10,065 sentences labelled by the annotators in the test set. 
Since all the sentences in the test set are checked by the annotators, the new test set does not have the FP and FN problem which are often occurred in the DS-constructed data.

\subsection{Data Post-Processing}
There are some relations that do not occur both in the train and test sets, and we decide to convert the labels of these relations into NA. 
After that, we continue to process the data by two steps: 
(1) if the number of instances for a relation is less than 100 in the train set, the relation will be converted into NA.
(2) If there are no instances labelled as ``Yes'' by the annotators for a relation in the test set, the relation will be converted into NA.
Finally, there are 550,720 instances (357,196 bags) in the NA set, 9,955 instances (3,548 bags) in the test set, and 21 relations (excluding NA) are kept in NYT-H.
\begin{figure*}[t]
    \centering    
    \begin{subfigure}[t]{0.3\textwidth}
        \centering
        \includegraphics[width=\textwidth]{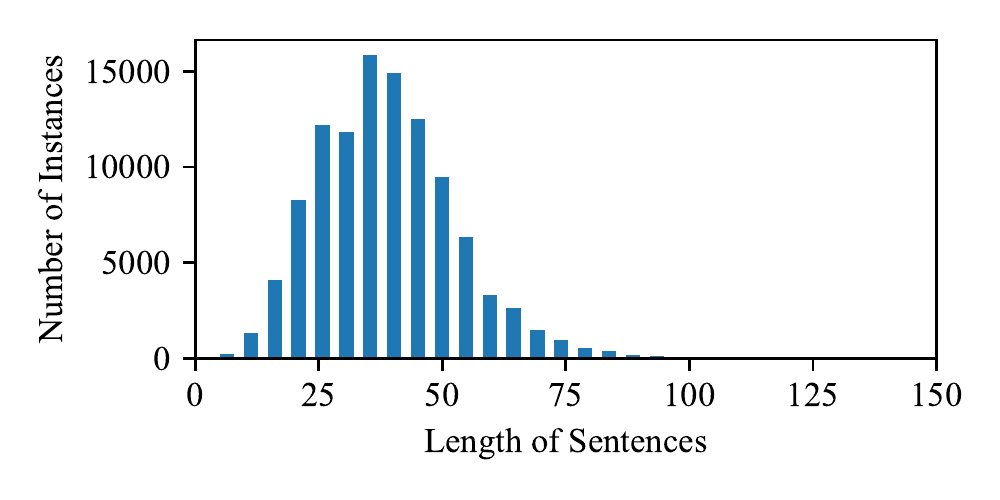}
        \caption{Train Set: Distribution of Sentence Length}
        \label{fig:trainsentlen}
    \end{subfigure}
    \hfill
    \begin{subfigure}[t]{0.3\textwidth}
        \centering
        \includegraphics[width=\textwidth]{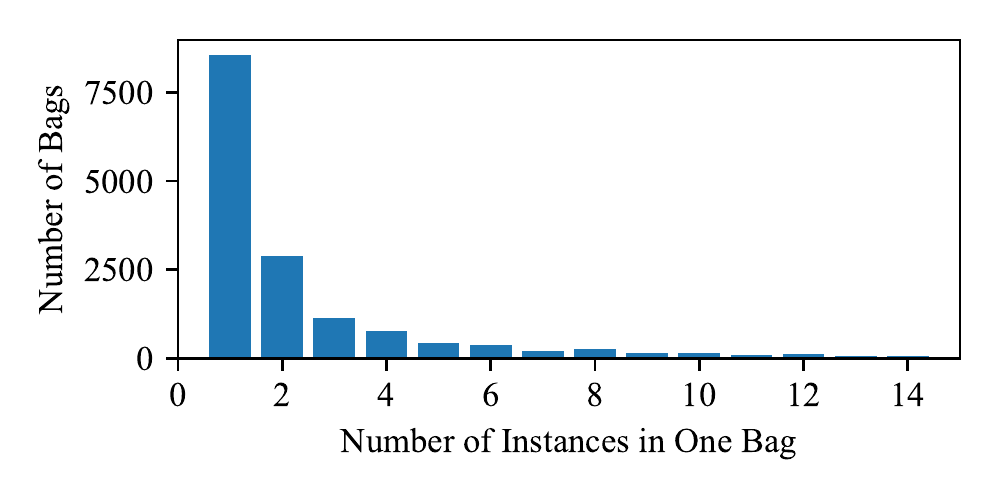}
        \caption{Train Set: Number of Instances in Bags}
        \label{fig:traininsinbags}
    \end{subfigure}
    \hfill
    \begin{subfigure}[t]{0.3\textwidth}
        \centering
        \includegraphics[width=\textwidth]{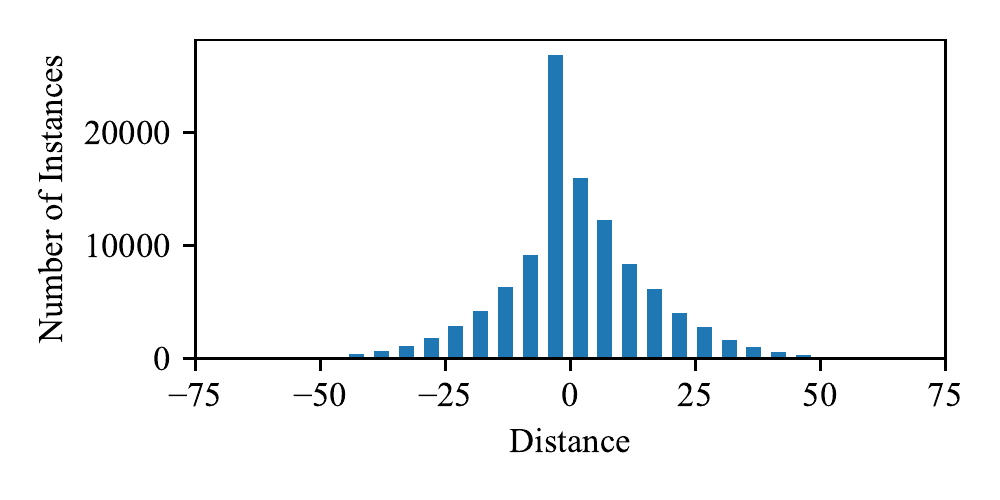}
        \caption{Train Set: Distance Between Entities}
        \label{fig:trainentdist}
    \end{subfigure}
    \newline
    
    \begin{subfigure}[t]{0.3\textwidth}
        \centering
        \includegraphics[width=\textwidth]{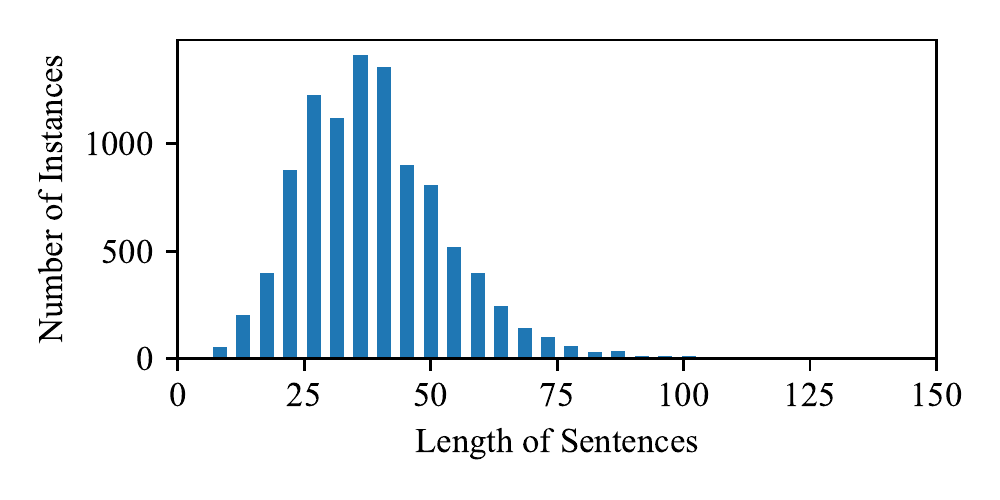}
        \caption{Test Set: Distribution of Sentence Length}
        \label{fig:testsentlen}
    \end{subfigure}
    \hfill
    \begin{subfigure}[t]{0.3\textwidth}
        \centering
        \includegraphics[width=\textwidth]{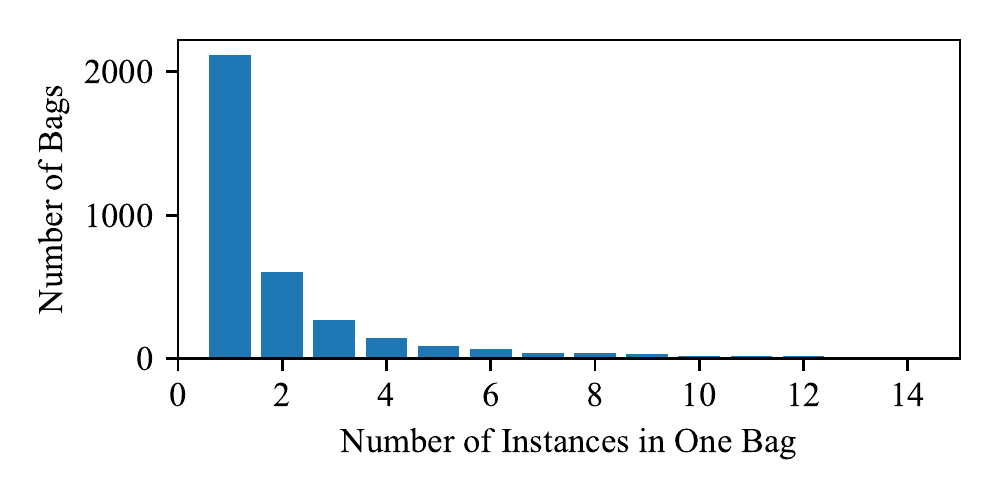}
        \caption{Test Set: Number of Instances in Bags}
        \label{fig:testinsinbags}
    \end{subfigure}
    \hfill
    \begin{subfigure}[t]{0.3\textwidth}
        \centering
        \includegraphics[width=\textwidth]{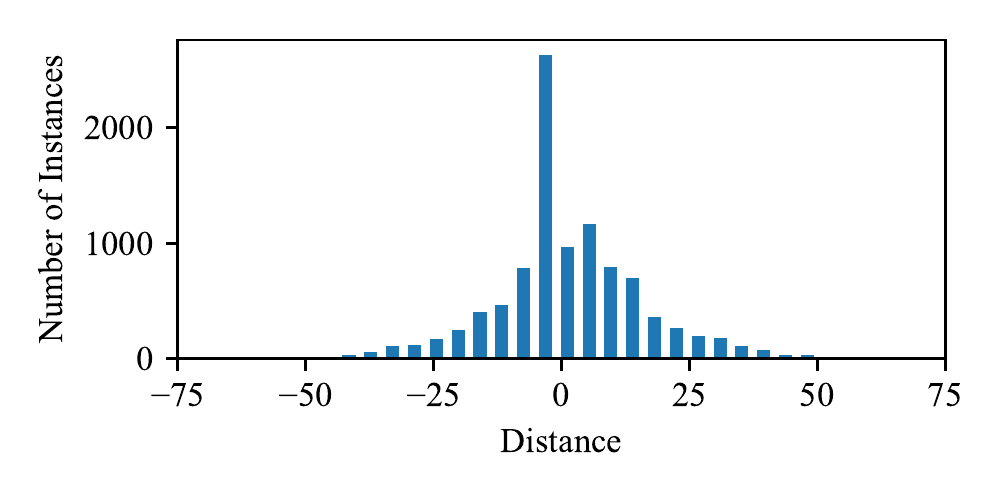}
        \caption{Test Set: Distance Between Entities}
        \label{fig:testentdist}
    \end{subfigure}
    \newline
    
    \begin{subfigure}[t]{0.3\textwidth}
        \centering
        \includegraphics[width=\textwidth]{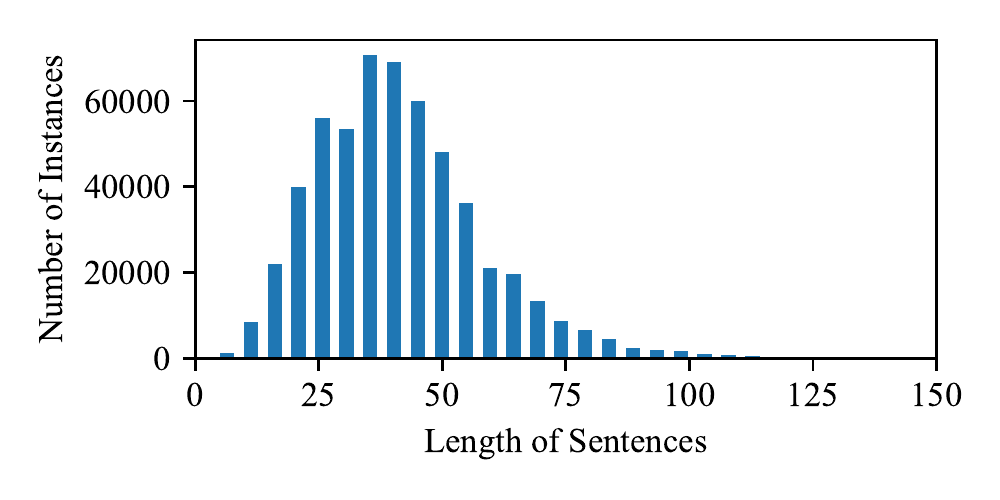}
        \caption{NA Set: Distribution of Sentence Length}
        \label{fig:nasentlen}
    \end{subfigure}
    \hfill
    \begin{subfigure}[t]{0.3\textwidth}
        \centering
        \includegraphics[width=\textwidth]{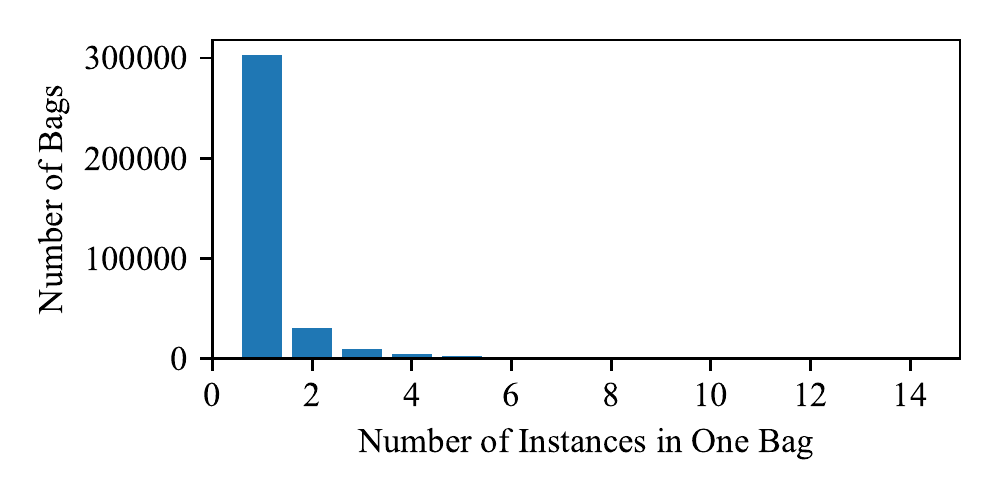}
        \caption{NA Set: Number of Instances in Bags}
        \label{fig:nainsinbags}
    \end{subfigure}
    \hfill
    \begin{subfigure}[t]{0.3\textwidth}
        \centering
        \includegraphics[width=\textwidth]{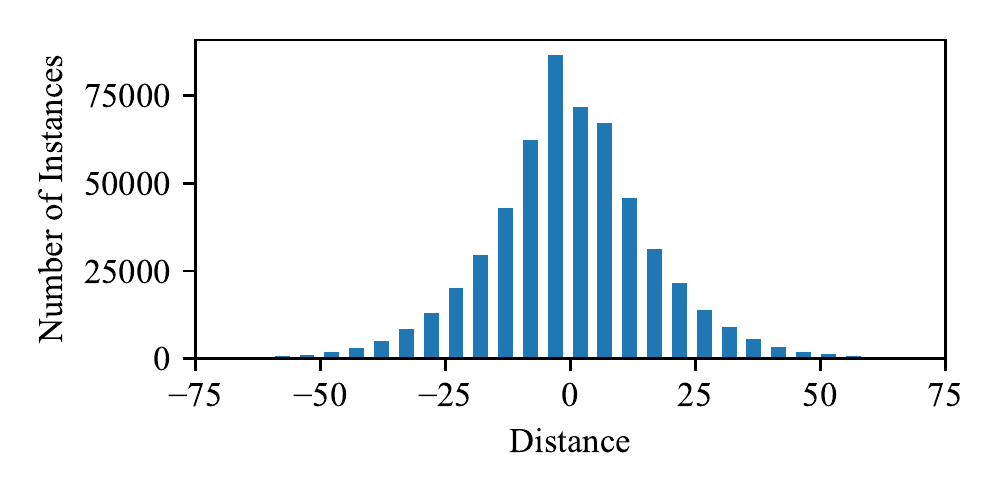}
        \caption{NA Set: Distance Between Entities}
        \label{fig:naentdist}
    \end{subfigure}

    \caption{Data statistics. The lengths of sentences are counted in word level. All the rectangles in bar and histogram plots are center aligned to the corresponding x coordinates. The distance between head entity and tail entity is obtained by subtracting the position of the head entity from the position of the tail entity.}
    \label{fig:datastats}
\end{figure*}

\subsection{Dataset Statistics}
Following the setting of MIL, the datasets are composed of bags. 
If there is at least one sentence in a bag manually labelled as ``Yes'', the DS-assigned relation will be the label of the bag. 
Thus, we can predict relations at bag-level and sentence-level.
In the test set, 5,202 out of 9,955 instances are annotated as ``Yes'', which also indicates the wrong label problem of distant supervision.
Besides, to make the dataset more consistent, relations are filtered to 21 (without NA relation), and all these relations are covered in both train and test set.
Figure \ref{fig:datastats} shows that the distributions of the train set are very close to those of the test set, which also indicates the consistency in NYT-H.

Although NYT-H is more consistent, it is still very challenging for building a relation extractor. 
As Table \ref{tab:dataset_scale} shows, it is an unbalanced dataset.
Besides, there are 3,548 bags in the test set, but Table \ref{tab:dataset_scale} shows the sum of test bag numbers is 3,735, which means there are some bags that have more than one relations.
In this condition, this dataset is also suitable for multi-instance multi-label research \cite{surdeanu2012miml,angeli2014combining,jiang2016relation,zhang2019capsule}.

\begin{table}[t]
    \centering
    \scalebox{0.8}{
        \begin{tabular}{l|cc|cc|cc}
            \toprule
            \multirow{2}{*}{Relation}&\multicolumn{2}{c|}{Train Set}&\multicolumn{4}{c}{Test Set}\\
             &\#Ins.&\#Bag&\#Ins.&\#Bag&\#Yes Ins.&\#Yes Bag\\
            \midrule
            /location/location/contains&51,059&7,488&4,475&1,603&2,754&1,219\\
            /people/person/nationality&8,575&2,298&1,128&523&731&320\\
            /location/country/capital&7,340&153&570&48&45&17\\
            /people/person/place\_lived&7,205&2,152&861&439&225&96\\
            /location/country/administrative\_divisions&6,301&351&481&98&284&59\\
            /location/administrative\_division/country&6,279&351&503&98&299&88\\
            /location/neighborhood/neighborhood\_of&5,629&462&176&83&104&55\\
            /business/person/company&5,602&1,361&591&278&468&230\\
            /people/person/place\_of\_birth&3,084&1,299&383&256&34&34\\
            /people/deceased\_person/place\_of\_death&1,914&670&193&117&54&41\\
            /business/company/founders&808&178&149&51&70&35\\
            /location/us\_state/capital&684&8&50&10&3&2\\
            /people/ethnicity/geographic\_distribution&531&26&137&5&17&4\\
            /people/person/children&479&205&57&42&45&35\\
            /business/company/place\_founded&464&162&57&33&21&17\\
            /business/company/major\_shareholders&291&27&47&9&3&2\\
            /sports/sports\_team\_location/teams&208&71&14&7&2&2\\
            /sports/sports\_team/location&204&71&28&13&15&8\\
            /people/ethnicity/people&150&54&5&5&3&3\\
            /people/person/ethnicity&145&54&22&7&9&4\\
            /people/person/religion&141&74&28&10&16&6\\
            \midrule
            SUM&107,093&17,515&9,955&3,735&5,202&2,277\\
            \bottomrule
        \end{tabular}
    }
    \caption{Statistics information of NYT-H. ``\#Yes Ins.'' and ``\#Yes Bag'' mean the number of instances and bags that are annotated as ``Yes''.}
    \label{tab:dataset_scale}
\end{table}

\subsection{Dataset Comparison}
Here, we compare our newly built data NYT-H with the previous datasets for relation extraction. 
The datasets can be divided into fully-annotated and distantly-supervised categories based on the constructing methods. 

\textbf{Fully-Annotated Datasets:} SemEval2010 Task8 \cite{hendrickx2009semeval}, ACE05 \cite{ace2005} and TACRED \cite{zhang2017position} are fully annotated.
As Table \ref{tab:dataset_comparison} shows, the scale of fully human annotated datasets are often small, both in the train and the test sets.
Among them, TACRED is the largest dataset, but only has 3,325 non-NA instances in the test set.
We can easily find that NYT-H has about three times as many non-NA test instances as TACRED.


\textbf{Distantly-Supervised Datasets:} Wiki-KBP \cite{ling2012fine}, GDS \cite{jat2018improving} and NYT10 \cite{riedel2010modeling} are all constructed by distant supervision.
NYT-Filtered \cite{zeng2015pcnn} and NYT-manual \cite{hoffmann2011multir} are generated from NYT10 with some predefined filtering rules. 
Among them, GDS, Wiki-KBP, and NYT-manual have human-annotated sentences in the test data like ours. 
This indicates that the researchers also realise the inaccurate evaluation problem in DS relation extraction. 
However, due to the cost, these human-annotated test sets are very small. 
Thus, the inaccurate evaluation problem is far from resolved. 
Therefore, in this paper we annotate more instances in the test set.

Besides, these datasets are facing an inconsistency problem, where relations in the train and the test sets are not fully overlapped.
Table \ref{tab:dataset_rel_comparison} shows that the number of relations in train and test sets are not the same in Wiki-KBP, NYT-Manual and NYT10. 
We believe that with NYT-H we can perform more accurate and consistent evaluation than ever.


\begin{table*}[t]
    \centering
    \scalebox{0.75}{
        \begin{tabular}{c|c|cccccccccccc}
            \toprule
             Type&Dataset Name&\#Ins.&\#Ent. Pair&\#Triple&\#Rel.&\#Ent.&\#Sent.&MA Test Set?&\makecell{\#Ins. in\\Test Set}&\makecell{\#Ins. in Test\\Set w/o NA} \\
             \midrule
             \multirow{3}{*}{MA}&ACE05-English&7,120&5,530&5,600&6&2,999&2,294&N.A.$\diamondsuit$&N.A.&N.A.\\
             &SemEval-2010 Task 8&10,717&10,233&10,281&19&7,858&10,674&Yes&2,717&2,717\\
             &TACRED&106,264&64,796&68,586&42&29,943&53,791&Yes&15,509&3,325\\
             \midrule
             \multirow{6}{*}{DS}&NYT10&742,748&375,914&377,495&58&69,063&320,711&No&172,448&6,444\\
             &NYT-Filtered&265,357&159,300&186,277&28&38,939&103,192&No&152,416&31,644\\
             &GDS&18,824&10,822&10,827&5&15,309&18,824&Partly$\heartsuit$&5,663&3,922\\
             &Wiki-KBP&153,966&131,534&133,050&13&40,415&23,884&Yes&2,209&316\\
             &NYT-Manual&376,733&203,340&204,835&25&53,047&210,325&Yes&3,880&410\\
             &NYT-H&667,806&375,829&377,393&22&69,063&320,668&Yes&9,955&9,955\\
             \bottomrule
        \end{tabular}
    }
    \caption{Widely-used English datasets for binary relation extraction (MA: manually annotated, DS: distantly supervised, Ins.: instance, Ent.: entity, Rel.: relation, Sent.: Sentence). $\diamondsuit$: ACE05 dataset does not have an available official test set. $\heartsuit$: GDS ensures the at-least-one assumption, but does not provide the detailed number of annotations, so the manually-annotated number ranges from 3,247 to 5,663 according to the number of bags in the test set.}
    \label{tab:dataset_comparison}
\end{table*}

\begin{table*}[h]
    \centering
    \scalebox{0.8}{
        \begin{tabular}{c|cccc}
            \toprule
            Dataset Name&\#Rel. in Train Set&\#Rel. in Test Set&\#Rel. in Both Sets&\#Total Rel.\\
            \midrule
            Wiki-KBP&7&13&7&13\\
            NYT-Manual&25&13&13&25\\
            NYT10&56&32&30&58\\
            NYT-H&22&22&22&22\\
            \bottomrule
        \end{tabular}
    }
    \caption{Statistics information of overlapped relations (Rel. means relations).}
    \label{tab:dataset_rel_comparison}
\end{table*}
\section{Evaluation Strategy}
\label{section:evaluation}
We expect that NYT-H can serve as a benchmark on the task of distantly-supervised relation extraction. In this section, we design three different tracks for evaluation on NYT-H at bag-level and sentence-level.

\subsection{Bag2Bag Track}
The Bag2Bag track is to evaluate the systems which are trained at bag-level and tested at bag-level. This setting is widely used for evaluating the MIL methods on the task of distantly-supervised relation extraction \cite{hoffmann2011multir,surdeanu2012miml,zeng2015pcnn,lin2016pcnnatt}. Thus, we also follow this setting of the previous studies. The difference is that the NYT-H data provides human-annotated labels in the test data.

We report the system quality by macro-averaged precision, recall, and $F_1$ scores.
The precision, recall and $F_1$ scores for relation $r_i$ in each system are calculated by the following equations:
\begin{equation}
\label{equation:pr}
\begin{split}
P^i   = \frac{N^i_{r}}{N^i_{sys}}, R^i   = \frac{N^i_{r}}{N^i_{data}}, F^i_1  = \frac{2 P^i  R^i}{P^i+R^i} \\
\end{split}
\end{equation}
where $N^i_r$ is the number of bags that the system correctly judges, $N^i_{sys}$ is the number of bags that the system predicts as relation $r_i$, and $N^i_{data}$ is the number of bags that the annotators label as relation $r_i$. 
Finally, the macro $F_1$ score of the system can be computed as follows: 
\begin{equation}
\label{equation:f1}
\begin{split}
    P = \frac{1}{m}\sum_{i=1}^{m} {P^i}, R = \frac{1}{m}\sum_{i=1}^{m}{R^i}, F_1  = \frac{1}{m}\sum_{i=1}^{m} {F^i_1}
\end{split}
\end{equation}
where $m$ denotes the number of relations. 

\subsection{Sent2Sent Track}
The Sent2Sent track is to evaluate the systems which are trained at sentence-level and tested at sentence-level. Researchers often utilize this strategy to the experiments on the fully human-annotated data.

Some studies use micro-averaged scores as the metrics in the Sent2Sent track \cite{zhang-etal-2017-end,sun-etal-2019-joint}, while
we report the systems' macro-averaged precision, recall, and $F_1$ scores to make relations equally weighted.
The scores are calculated using Equation \eqref{equation:pr} and \eqref{equation:f1} as in the Bag2Bag track. The difference is the definition of the numbers, where $N^i_r$ is the number of sentences that the system correctly judges, $N^i_{sys}$ is the number of sentences that the system predicts as relation $r_i$, and $N^i_{data}$ is the number of sentences that the annotators label as relation $r_i$.

\subsection{Bag2Sent Track}
The Bag2Sent track is to evaluate the systems which are trained at bag-level and tested at sentence-level. NYT-H offers a compromise of labelled data between distant supervision and manual annotation. Thus, we propose the Bag2Sent track, where we train models on distantly-supervised data at bag-level and evaluate the models on manually annotated data at sentence-level. The evaluation scores are calculated in the same way as the method in Sent2Sent track.

\section{Experiments}
In this section, we present the experimental results of several widely-used RE systems on NYT-H.

\subsection{Experimental Settings}
We randomly select 10\% instances from the train set as a development set to help the model selection. We use 5 different random seeds and take the mean scores as final results.
In addition, NA instances are sub-sampled from the NA set with the same scale of the train set.
We use 50 dimensional GloVe\footnote{\url{https://nlp.stanford.edu/projects/glove/
}} vectors as the system input, and other parameter settings are listed in Table \ref{tab:parameter_settings}.

\begin{table}[h]
    \centering
    \scalebox{0.8}{
        \begin{tabular}{ccccc}
             \toprule
             Epoch & Batch Size & Dropout Rate & Learning Rate&Dimension of Position Features\\
             \midrule
             50&64&0.5&$1\times 10^{-3}$&5\\
             \bottomrule
        \end{tabular}
    }
    \caption{Parameter settings in experiments.}
    \label{tab:parameter_settings}
\end{table}

\subsection{Comparison Systems}
In this section, we list the systems used in our experiments. There are two groups of systems: training at sentence-level and bag-level. 

\textbf{Sentence-Level:} Convolutional Neural Network (CNN) \cite{zeng2014cnn} is able to capture local features effectively. Piecewise Convolutional Neural Network (PCNN) \cite{zeng2015pcnn} and Classification by Ranking Convolutional Neural Network (CR-CNN) \cite{santos2015crcnn} are variations of CNN, where PCNN can retain richer information and CR-CNN has a strong ability to reduce NA impact. ATT-BLSTM \cite{zhou2016attblstm} is an RNN-based model which can capture global features. These systems are all trained at sentence-level.

\textbf{Bag-Level:} CNN and PCNN can be further combined with ONE \cite{zeng2015pcnn}\footnote{\url{https://github.com/smilelhh/ds_pcnns}} and ATT \cite{lin2016pcnnatt}\footnote{\url{https://github.com/thunlp/OpenNRE}} to train at bag-level, where ONE selects the most valuable sentence per bag and ATT generates a weighted sentence representation via attention mechanism for each bag.

\subsection{Evaluation Results}

Two types of evaluation results are presented here. The first one is \textbf{DS} labels as \textbf{G}round \textbf{T}ruths (DSGT), which assumes that all the DS labels are ground truths. 
This is a common setting used on distantly-supervised relation extraction, but may result in incorrect evaluation during testing.
The second one is the \textbf{M}anual \textbf{A}nnotation as \textbf{G}round \textbf{T}ruth (MAGT), where the metrics are calculated on the same evaluation set, but uses the human-annotated labels as truths.

\begin{table*}[t]
    \centering
    \scalebox{0.8}{
        \begin{tabular}{c|c|c|ccc|ccc}
            \toprule
             \multirow{2}{3em}{Tracks} &
             \multirow{2}{3em}{Models}& \multirow{2}{2em}{AUC} & \multicolumn{3}{c|}{DSGT (\%)} & \multicolumn{3}{c}{MAGT (\%)}\\
             &&  & Precision & Recall & F1 & Precision & Recall & F1\\
             \midrule
             \midrule
             \multirow{4}{4em}{Sent2Sent} & CNN &  - & 71.560 & 47.190 & 54.707 & 41.656 & 47.291 & 38.989\\
             & CR-CNN & - & 72.016 & \textbf{55.796} & \textbf{60.953} & 43.961 & 58.893 & 45.060\\
             & PCNN & - & \textbf{72.194} & 50.791 & 57.687 & 44.667 & 54.703 & 44.011\\
             & ATT-BLSTM & - & 71.972 & 55.313 & 60.165 & \textbf{45.336} & \textbf{60.004} & \textbf{45.928}\\
             \midrule
             \multirow{4}{4em}{Bag2Sent} & CNN+ONE& - & 64.970 & 24.777 & 32.711 & 48.501 & 28.096 & 31.695 \\
             & CNN+ATT & - & \textbf{65.996} & 22.729 & 30.976 & 50.334 & 26.239 & 30.488 \\
             & PCNN+ONE & - & 64.020 & \textbf{26.362} & \textbf{33.893} & \textbf{51.787} & \textbf{32.240} & \textbf{34.981} \\
             & PCNN+ATT & - & 63.542 & 24.388 & 31.913 & 48.728 & 28.334 & 32.367  \\
             \midrule
             \multirow{4}{4em}{Bag2Bag} & CNN+ONE& 0.671 & \textbf{66.823} & \textbf{37.191} & \textbf{45.325} & 43.478 & 45.078 & 39.539 \\
             & CNN+ATT & 0.690 & 57.942 & 23.823 & 31.660 & \textbf{50.632} & 21.792 & 26.433\\
             & PCNN+ONE & 0.681 & 63.096 & 37.010 & 44.299 & 45.586 & \textbf{47.206} & \textbf{41.843} \\
             & PCNN+ATT & \textbf{0.699} & 58.269 & 27.124 & 34.879 & 48.121 & 24.952 & 28.805 \\
             \bottomrule
        \end{tabular}
    }
    \caption{Experiment Results. AUC scores are calculated via micro averaged method provided by OpenNRE.}
    \label{tab:exp_results}
\end{table*}

\begin{table*}[]
    \centering
    \scalebox{0.8}{
        \begin{tabular}{c|cccccc}
            \toprule
            Model&P@50&P@100&P@300&P@500&P@1000&P@2000\\
            \midrule
            CNN+ONE&0.924&0.900&0.869&0.854&0.822&0.745\\
            CNN+ATT&0.920&0.914&0.889&0.859&0.818&0.746\\
            PCNN+ONE&0.928&0.91&0.872&0.862&0.828&0.756\\
            PCNN+ATT&\textbf{0.940}&\textbf{0.918}&\textbf{0.909}&\textbf{0.880}&\textbf{0.834}&\textbf{0.759}\\
            \bottomrule
        \end{tabular}
    }
    \caption{Precision@K results of Bag2Bag track.}
    \label{tab:exp_pk_results}
\end{table*}

\begin{figure}[ht]
    \centering
    \includegraphics[scale=0.5]{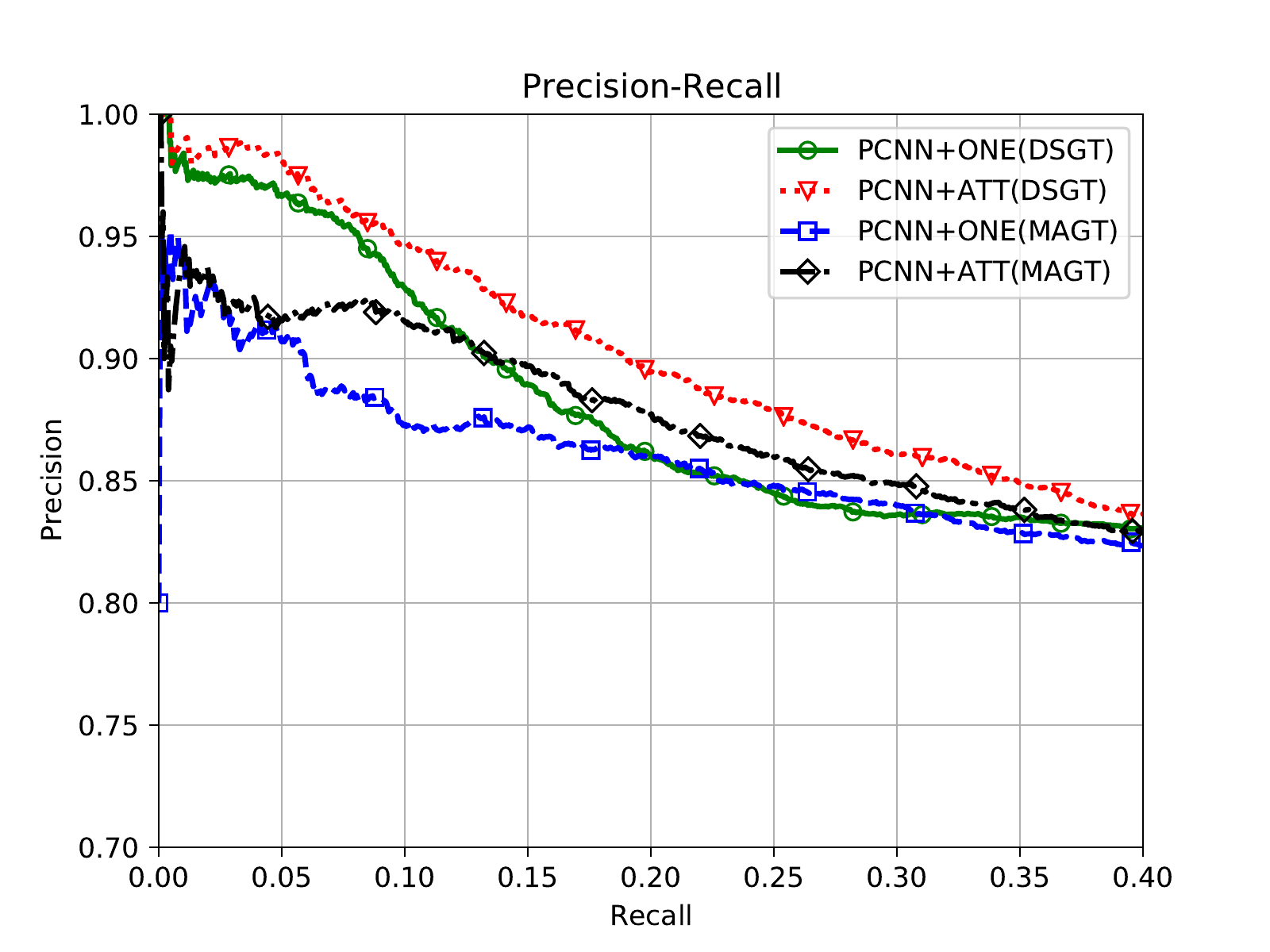}
    \caption{Precision-Recall curves in Bag2Bag track.}
    \label{fig:prc_merge}
\end{figure}

From the $F_1$ results in Table \ref{tab:exp_results}, we can observe the following facts:

\begin{itemize}
    \item In Sent2Sent track, there are obvious downward $F_1$ gaps (about 15\%) from DSGT to MAGT. This indicates that the $F_1$ scores of DSGT are not accurate and the sentence-level systems are sensitive to the DS noises. As for DSGT, CR-CNN achieves the best performance, while it gets the second on MAGT. 

    \item In Bag2Sent track, all the bag-level models show a much better ability to deal with the DS noises, since the gap is not that much. Although the MIL strategy does great on noise deduction during training, the final MAGT results of the bag-level models are worse than sentence-level models. 

    \item In Bag2Bag track, the ranking order is changed because of the DS noises. 
The results of AUC show PCNN+ATT gets the best performance, and the best model on DSGT $F_1$ is CNN+ONE, while the order of MAGT is PCNN+ONE $>$ CNN+ONE $>$ PCNN+ATT $>$ CNN+ATT. 
These facts indicate that the results of AUC and $F_1$ in the DS test set are not reliable to draw objective conclusions.
Bag2Bag track is the most common setting in DS relation extraction due to the bag assumption.
The Precision-Recall curves are widely-used evaluation metric in Bag2Bag track, and curves in Figure \ref{fig:prc_merge} also indicate the evaluation problem.
There are conspicuous gaps between DSGT curves and MAGT curves. 

\end{itemize}

In summary, we find: 
(1) $F_1$ scores on MAGT are much lower than the ones on DSGT in most cases, and the gaps can also be found in Figure \ref{fig:prc_merge}. 
(2) The ranking lists of the systems are different in many cases between two test settings, especially in Bag2Bag track which is the most used setting in DS relation extraction. 
The evaluation on the DS-generated test data produces incorrect scores that may lead to misjudgement. 
Thus, we believe that the accurate evaluation can boost the research on distantly-supervised relation extraction.

\subsection{Relation Coverage in Precision@K Evaluation}
As we have mentioned in Section \ref{related_work}, P@K is another widely used measure in DS relation extraction, which is mainly evaluated by manually check for each system. Since we have the human-annotated test data, we can automatically obtain the scores of P@K, even `K' is much larger than ever. Table \ref{tab:exp_pk_results} shows the P@K results where `K' ranges from 50 to 2,000.

\begin{figure*}[h]
    \centering
    \begin{subfigure}[t]{0.4\textwidth}
        \centering
        \includegraphics[width=\textwidth]{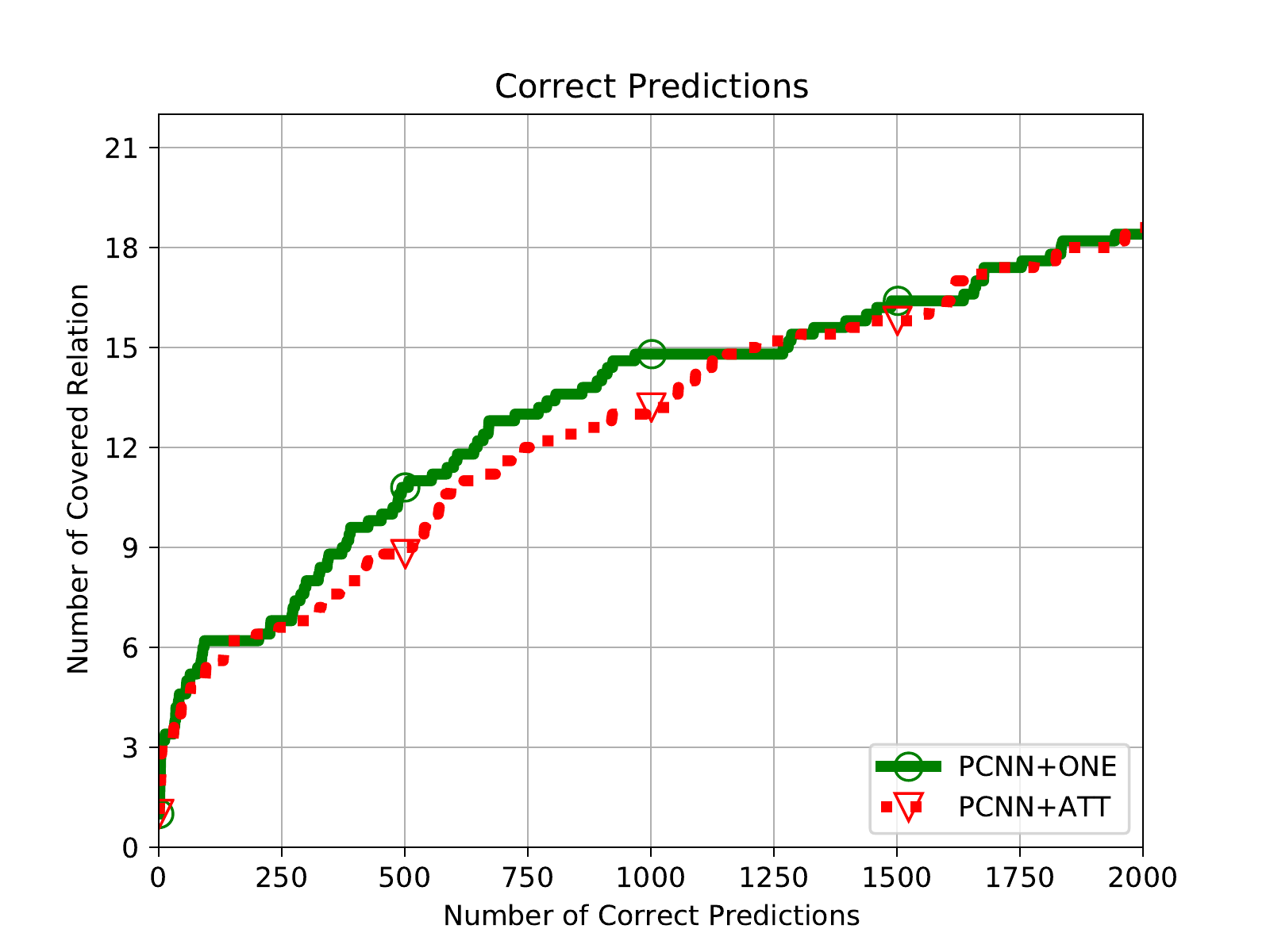}
        \caption{Relation Coverage for Correct Predictions}
        \label{fig:num_of_covered_rel_correct}
    \end{subfigure}
    \begin{subfigure}[t]{0.4\textwidth}
        \centering
        \includegraphics[width=\textwidth]{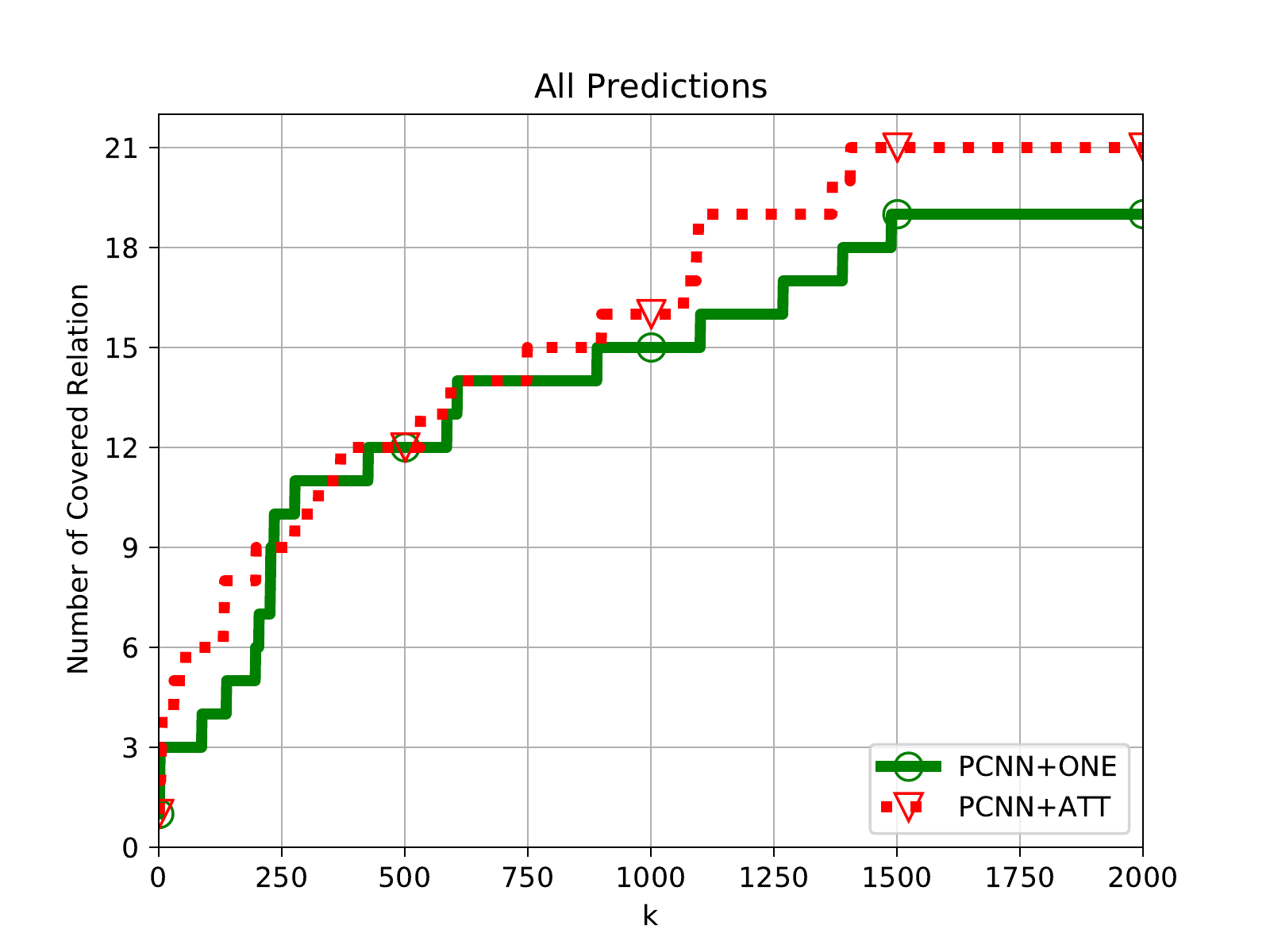}
        \caption{Relation Coverage for All Predictions}
        \label{fig:num_of_covered_rel_all}
    \end{subfigure}
    \caption[]{Relation coverage when K changes}
    \label{fig:num_of_covered_rel}
\end{figure*}

We further check relation coverage when performing P@K evaluation. Figure \ref{fig:num_of_covered_rel_correct} shows the results of relation coverage for correct prediction, while Figure \ref{fig:num_of_covered_rel_all} shows the results for all predictions. 
From the figure, we can find that PCNN+ONE could not cover all the 21 non-NA relations in the top 2,000 results, while PCNN+ATT covered all the relations until `K' equals to 1,405.

As we have discussed in section \ref{related_work}, the `K' value selection is crucial to P@K evaluation. 
P@K is a good way to measure predictions especially when the scale of predicted results are too large to be fully annotated. However, relations cannot be fully covered if the `K' is too small.
Figure \ref{fig:num_of_covered_rel_all} shows that only 5 relations are covered in PCNN+ATT when `K' is 50, while the number of the whole relation candidate set is 21. 
As Table \ref{tab:exp_results} shows, none of the $F_1$ scores in the three tracks exceed 50\%, which is far below our expectations.
But the P@K results give us an illusion that the reuslts have reached the ceiling, which may lead to an evaluation bias.


\section{Conclusion}
In this paper, we present the NYT-H dataset, an enhanced distantly-supervised dataset with human annotation, where we use DS-generated training data and human-annotated test data. 
The NYT-H dataset can resolve the inaccurate evaluation problem caused by the assumption of distant supervision. 
To compare the performance of the previous systems, we design three evaluation tracks, which are Sent2Sent, Bag2Sent and Bag2Bag. 
We conduct experiments on the newly built NYT-H data with widely used baseline systems for comparisons. 
All the evaluation scripts and data resources are available in \url{https://github.com/Spico197/NYT-H}.

\section{Acknowledgement}
The research work is supported by the National Natural Science Foundation
of China (Grant Nos. 61525205 and 61936010).
This
work is partially supported by the joint research project
of Alibaba and Soochow University.
Corresponding author
is Xiabing Zhou.
We would also thank the anonymous
reviewers for their detailed comments, which have helped us
to improve the quality of this work.

\bibliographystyle{coling}
\bibliography{references}

\end{document}